\documentclass[10pt,twocolumn,letterpaper]{article}

\usepackage[pagenumbers]{cvpr}

\usepackage[utf8]{inputenc} 
\usepackage[T1]{fontenc}    
\usepackage{inconsolata}    
\usepackage{nicefrac}       
\usepackage{microtype}      

\usepackage{multirow}
\usepackage{tabularx}
\usepackage[frozencache,cachedir=minted-cache]{minted}
\usepackage{siunitx}

\definecolor{cvprblue}{rgb}{0.21,0.49,0.74}
\usepackage[pagebackref,breaklinks,colorlinks,allcolors=cvprblue]{hyperref}

\def\confName{CVPR}
\def\confYear{2027}

\title{Large-Scale Gaussian Splatting with a Multi-GPU PyTorch Backend}

\author{%
  Matthew Cong\textsuperscript{1} \qquad
  Francis Williams\textsuperscript{1} \qquad
  Jonathan Swartz\textsuperscript{1} \\[2pt]
  Mark Harris\textsuperscript{1} \qquad
  Sanja Fidler\textsuperscript{1,2,3} \qquad
  Ken Museth\textsuperscript{1} \\[6pt]
  {\normalsize
   \textsuperscript{1}NVIDIA \qquad
   \textsuperscript{2}University of Toronto \qquad
   \textsuperscript{3}Vector Institute}%
}

\begin{document}

\maketitle

\begin{abstract}
Gaussian splatting methods have become increasingly popular for neural reconstruction of the real world.
However, they are often limited in scale and resolution due to compute and memory constraints.
We present a multi-GPU Gaussian splatting approach that scales reconstruction to higher resolutions and larger scenes while abstracting away the code complexity typically associated with distributing a model.
We demonstrate city-scale reconstructions with street-level detail consisting of over 1 billion Gaussian splats.
To accomplish this, we propose a PyTorch backend that distributes parameters and operators across GPUs via CUDA unified memory, with optional NVLink support.
Because distribution occurs at the operator level, the model code requires no explicit cross-device communication.
More broadly, the backend exposes multiple GPUs as an aggregate asynchronous PyTorch device and supports a broad set of PyTorch operators enabling the distribution of general PyTorch workloads.
\end{abstract}

\begin{figure*}[t]
    \centering
    \includegraphics[width=0.66\linewidth]{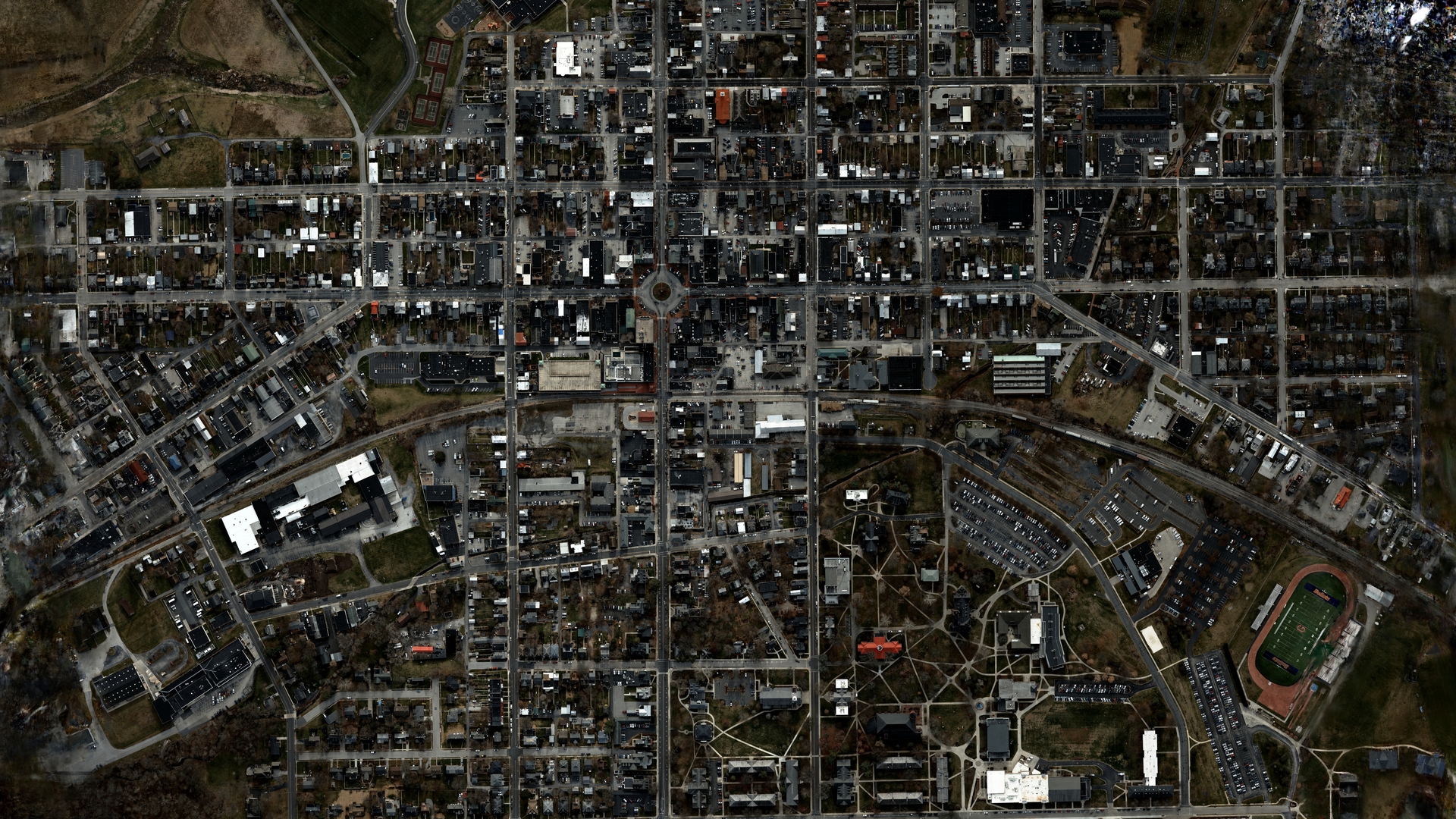}
    \includegraphics[width=0.33\linewidth]{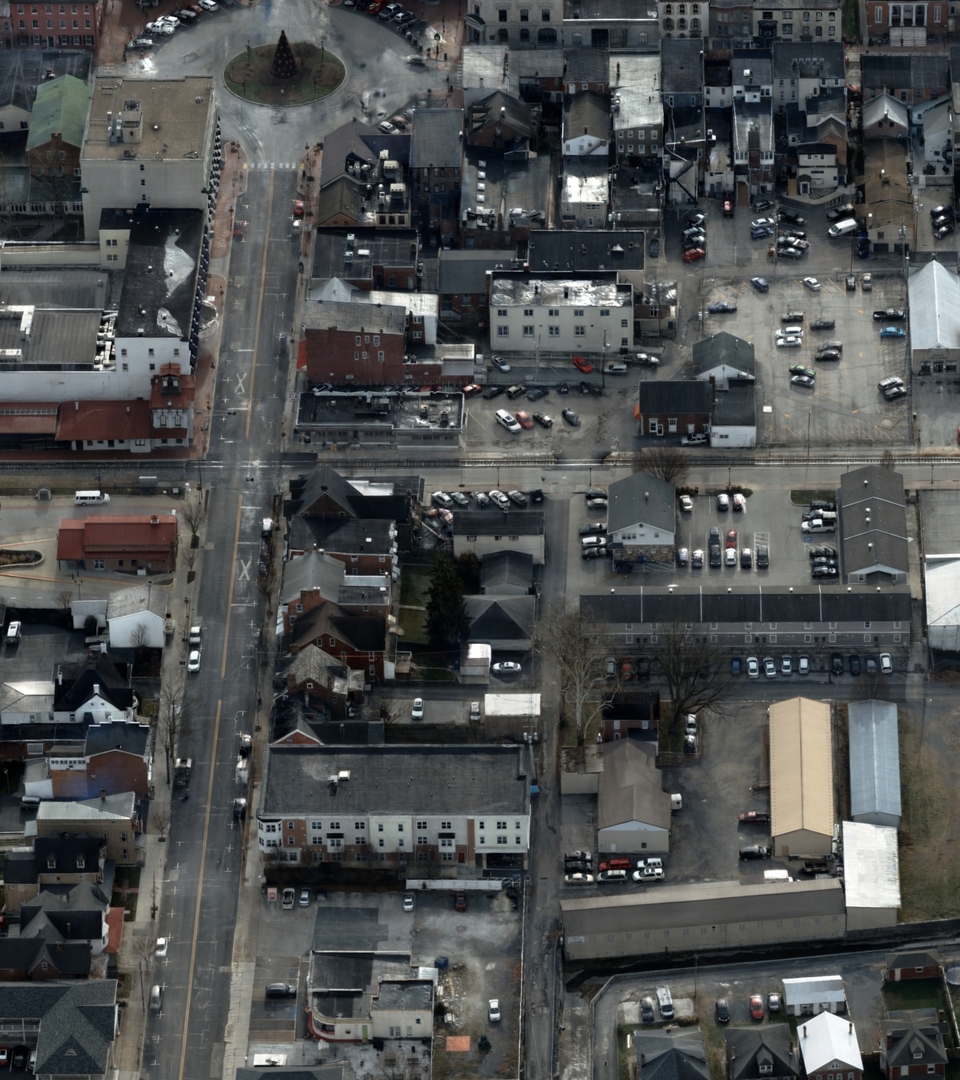}
    \caption{
        A large-scale high-resolution reconstruction of over 268 million Gaussian splats trained using our multi-GPU Gaussian splatting approach with a batch size of 1.
        The dataset consists of 153 aerial images at $7088 \times 5314$ resolution covering 13 square kilometers, with camera poses initialized via COLMAP \cite{schoenberger2016sfm}.
        Holding out 1 in every 10 images for evaluation, we obtain a PSNR of 27.450 and an SSIM of 0.8734 after training for 400 epochs on 8 A100 GPUs with a wall-clock time of 3.086 hours.
        For comparison, the implementation of Grendel-GS \cite{zhao2025on} in gsplat \cite{ye2025gsplat} runs out of memory at 132 million Gaussians on 8 A100 GPUs with its minimum supported batch size ($B=8$) for this GPU count, where each view's projected parameters must lie entirely on a single GPU.
        \emph{Left:}~overhead view of the town and surrounding fields.
        \emph{Right:}~close-up showing vehicles, fa\c{c}ades, and road markings.
        Gettysburg dataset courtesy of the Army Geospatial Lab.
        \label{figure:gettysburg_reconstruction}}
\end{figure*}

\vspace{-0.3in}

\section{Introduction}

Gaussian splatting \cite{kerbl2023gaussian} is a rapidly evolving area of research with applications including three-dimensional scene reconstruction, novel view synthesis, promptable segmentation \cite{kirillov2023segment}, and physical simulation \cite{xie2023physgaussian}.
Given its advantages over prior NeRF-based approaches \cite{mildenhall2021nerf}, it has drawn significant interest from academia and industry, motivating several open-source frameworks \cite{williams2024fvdb, ye2025gsplat}.
It has also spawned a family of techniques, with extensions such as 3DGUT \cite{wu20253dgut}, RPC-GS \cite{wagner2026rpcgsgaussiansplattingnative}, and Gaussian Wrapping \cite{gomez2026blobsspokeshighfidelitysurface} supporting different camera models, rendering pipelines, and surface reconstruction methods.
However, most of these frameworks are constrained by the compute and memory of a single GPU, limiting reconstruction of large-scale scenes from high-resolution inputs.
Adapting each implementation for multi-GPU execution requires specialized code to manage tensor residency, workload partitioning, synchronization, and communication.

We seek a more general approach to multi-GPU execution, designed to accommodate both Gaussian splatting and general PyTorch workloads.
To this end, we introduce \texttt{torch-dgx}, a multi-GPU PyTorch \cite{paszke2019pytorch} backend that distributes individual operators along their natural axes of parallelism, which can vary across operators and may even depend on the operator's inputs at runtime.
Because distribution is handled within operator implementations, model code does not need to explicitly manage cross-device communication or synchronization.
The backend uses CUDA unified memory to provide a unified virtual address space across GPUs, with optional NVLink support for high-throughput all-to-all communication.
We further propose a fully asynchronous allocator for unified-memory tensors to avoid the overhead of host-synchronous allocation calls.
The resulting backend supports a broad set of PyTorch operators enabling the distribution of general PyTorch workloads.

Building on this backend, we introduce a Gaussian splatting approach that supports both single-camera and batched training across multiple GPUs.
It distributes both Gaussian parameters and per-view rendering data across GPUs, allowing a single camera view to use their aggregate memory and compute.
Other Gaussian splatting variants can reuse the backend's distributed PyTorch operators when adapting their pipelines to multi-GPU execution.
Experiments demonstrate faster training than single-GPU execution with both increasing and constant batch sizes, while maintaining comparable reconstruction quality.
We also demonstrate high-resolution reconstructions at Gaussian counts beyond the memory capacity of comparable multi-GPU Gaussian splatting implementations (Figure~\ref{figure:gettysburg_reconstruction}).
By making aggregate GPU resources available to a single camera view, our approach enables city-scale reconstructions with street-level detail consisting of over 1 billion Gaussian splats.

\section{Related Work}

\subsection{Scalable Gaussian Splatting}

Prior efforts to scale Gaussian splatting have explored divide-and-conquer approaches, level-of-detail representations, acceleration structures, and multi-GPU execution.
Domain-decomposition approaches \cite{yuchen2024dogaussian, lin2024vastgaussian, liu2025citygaussian} approximate the full reconstruction problem by partitioning the scene into weakly coupled sections and reconstructing each section independently.
Since the sections are not truly independent, each is often expanded slightly to overlap with neighboring sections so that seams can be blended at the boundaries, incurring overhead proportional to the number of GPUs.
NeRF-XL \cite{li2024nerfxl} avoids this approximation by reformulating the underlying mathematical problem and coupling neighboring sections via cross-GPU communication during training.
Grendel-GS \cite{zhao2025on} also solves the full Gaussian splatting problem, using per-Gaussian and per-pixel parallelism to distribute training across GPUs.
Together with spatial locality and dynamic load balancing, this enables Grendel-GS to scale to over 40 million Gaussians across 16 GPUs at 4K resolution.
Hierarchical and level-of-detail approaches build acceleration structures over the Gaussian splats to speed up intersection and enable level-of-detail rendering \cite{kerbl2024hierarchical, lu2024scaffold, ren2025octree}, at the cost of additional compute and memory to traverse, build, and store the structure.
In contrast to these higher-level algorithmic improvements, LiteGS \cite{liao2025litegs} addresses bottlenecks for Gaussian splat training at the system level.
The higher-level algorithmic improvements are complementary to our multi-GPU approach and could be adopted alongside the distributed operators in \texttt{torch-dgx}.

\subsection{Distributed Training}

At scales of billions to trillions of parameters, modern neural networks routinely exceed the compute and memory of a single GPU and are distributed across multiple GPUs in several ways.
Data parallel approaches \cite{li2020pytorchddp} replicate the model across GPUs, train on distinct batches, and aggregate the resulting gradients and parameter updates.
Tensor parallel approaches such as GSPMD \cite{xu2021gspmd} shard parameters across GPUs, with each GPU operating on its subset and communicating with neighbors as needed.
Pipeline parallel approaches such as GPipe \cite{huang2019gpipe} partition the computation graph into stages distributed across GPUs, where the output of a stage on one GPU is pipelined to the next stage on another GPU.
Many systems combine these paradigms: Fully Sharded Data Parallel \cite{zhao2023pytorch} merges data and tensor parallelism by gathering and scattering parameters and gradients on a layer-by-layer basis during training, and the approach in \cite{narayanan2021efficient} combines pipeline, tensor, and data parallelism to train large-scale language models.

While these approaches are theoretically agnostic to the underlying model, in practice they are tightly coupled to the model, and adopting them typically requires significant changes to handle communication, sharding, and residency.
For example, PyTorch's DTensor framework \cite{pytorch2025dtensor} requires explicitly specifying sharding and replication for each tensor \emph{a priori}, making efficient distribution a nontrivial task that demands substantial PyTorch and parallel-programming expertise.
A complementary approach expands the memory available to a single GPU by using CUDA unified memory to spill PyTorch tensor storage into host memory \cite{choi2021implementing}.
Our approach instead aggregates GPU memory and distributes computation across multiple GPUs, with minimal host memory involvement.

\begin{table}[t]
    \centering
    \caption{
        \texttt{torch-dgx} \emph{Operator Benchmarks}: We report timings and relative speeds for several distributed operators on $N \in \{1, 2, 4, 8\}$ A100 GPUs connected with NVLink.
        Each operator is benchmarked on randomly initialized \texttt{float} tensors with $K \in \{2^{26}, 2^{28}, 2^{30}\}$ elements. For \texttt{addmm}, $K$ denotes the total entry count of the output matrix where the inputs have $\sqrt{K}$ rows and columns.
        Reported timings average 20 iterations after 10 warmup iterations.
        At smaller problem sizes on larger GPU counts, there is often insufficient work to saturate per-GPU compute and memory bandwidth.
        However, as input tensor size grows, our operators approach linear scaling.
        \label{table:microbenchmarks}}
    \scriptsize
    \setlength{\tabcolsep}{3pt}
    \begin{tabular*}{0.965\columnwidth}{@{\extracolsep{\fill}}l c cc cc cc@{}}
        \toprule
        & & \multicolumn{2}{c}{$K = 2^{26}$} & \multicolumn{2}{c}{$K = 2^{28}$} & \multicolumn{2}{c}{$K = 2^{30}$} \\
        \cmidrule(lr){3-4} \cmidrule(lr){5-6} \cmidrule(lr){7-8}
        Operator & GPUs & {\unit{\milli\second}} & {$\times$} & {\unit{\milli\second}} & {$\times$} & {\unit{\milli\second}} & {$\times$} \\
        \midrule
        \multirow{4}{*}{\texttt{fill}} & 1 & 0.1991 & 1.00 & 0.7815 & 1.00 & 3.1123 & 1.00 \\
        & 2 & 0.1171 & 1.70 & 0.4085 & 1.91 & 1.5743 & 1.98 \\
        & 4 & 0.0733 & 2.72 & 0.2188 & 3.57 & 0.8019 & 3.88 \\
        & 8 & 0.0771 & 2.58 & 0.1310 & 5.97 & 0.4224 & 7.37 \\
        \midrule
        \multirow{4}{*}{\texttt{add}} & 1 & 0.4849 & 1.00 & 1.9133 & 1.00 & 7.6206 & 1.00 \\
        & 2 & 0.2739 & 1.77 & 0.9887 & 1.94 & 3.8437 & 1.98 \\
        & 4 & 0.1629 & 2.98 & 0.5208 & 3.67 & 1.9505 & 3.91 \\
        & 8 & 0.1395 & 3.48 & 0.2989 & 6.40 & 1.0184 & 7.48 \\
        \midrule
        \multirow{4}{*}{\texttt{addmm}} & 1 & 58.13 & 1.00 & 466.39 & 1.00 & 3966.25 & 1.00 \\
        & 2 & 29.79 & 1.95 & 236.55 & 1.97 & 1854.76 & 2.14 \\
        & 4 & 16.46 & 3.53 & 122.84 & 3.80 & 949.42 & 4.18 \\
        & 8 & 11.18 & 5.20 & 73.91 & 6.31 & 528.32 & 7.51 \\
        \midrule
        \multirow{4}{*}{\texttt{sum}} & 1 & 0.1587 & 1.00 & 0.5763 & 1.00 & 2.2517 & 1.00 \\
        & 2 & 0.1245 & 1.27 & 0.3306 & 1.74 & 1.1693 & 1.93 \\
        & 4 & 0.1426 & 1.11 & 0.2066 & 2.79 & 0.6256 & 3.60 \\
        & 8 & 0.2401 & 0.66 & 0.2570 & 2.24 & 0.3732 & 6.03 \\
        \midrule
        \multirow{4}{*}{\texttt{cumsum}} & 1 & 0.5253 & 1.00 & 1.9625 & 1.00 & 7.7504 & 1.00 \\
        & 2 & 0.3354 & 1.57 & 1.0592 & 1.85 & 3.9475 & 1.96 \\
        & 4 & 0.2496 & 2.10 & 0.6023 & 3.26 & 2.0477 & 3.78 \\
        & 8 & 0.4621 & 1.14 & 0.4958 & 3.96 & 1.1295 & 6.86 \\
        \bottomrule
    \end{tabular*}
\end{table}

\section{Multi-GPU PyTorch Backend}\label{section:backend}

CUDA unified memory exposes the physical memory of multiple GPUs through a single virtual address space with physical residency managed at runtime by the driver.
NVLink, while not strictly required, further improves the performance of unified memory by providing high-throughput communication between GPUs.
Together, these provide a natural foundation for distributing large-scale models.
However, PyTorch does not natively support CUDA unified memory.

To this end, \texttt{torch-dgx} combines two key components: unified memory allocation and deallocation of PyTorch tensors, and multi-GPU implementations of PyTorch operators.
Both memory management and operator execution are asynchronous with respect to the host.
Unified memory provides shared access to tensor storage in a single virtual address space, which allows the \texttt{Tensor} struct to remain unchanged.
Meanwhile, each operator implementation chooses its work partition, coordinates execution across GPUs, and optimizes data locality.
For models composed of PyTorch operators, selecting \texttt{device="dgx"} distributes tensor allocation and operator execution across GPUs with minimal changes to model code.
Custom operators require multi-GPU implementations, as illustrated by the Gaussian splatting operators in Section~\ref{section:distributed_gaussian_splatting}.
A complete and self-contained training example appears in Listing~\ref{listing:code_sample} in the supplementary material.

\subsection{Tensor Memory Management}

Traditionally, the underlying storage for PyTorch GPU tensors is allocated on a single device via \texttt{cudaMalloc} with a pseudo-asynchronous block allocator overlaid for efficient memory reuse.
As a first attempt, one could simply replace \texttt{cudaMalloc} with \texttt{cudaMallocManaged} in order to allocate a tensor in unified memory accessible to all GPUs.
Although this approach is sufficient for correct behavior, in practice it exhibits severe performance degradation as PyTorch frequently allocates and deallocates tensors.
First, the PyTorch block allocator does not generalize to the multi-GPU case, and thus cannot provide memory reuse or pseudo-asynchronous allocation across devices.
Second, \texttt{cudaMallocManaged} and \texttt{cudaFree} are synchronous with respect to the host and involve relatively expensive system calls.
Thus, to allocate or deallocate memory, the host must synchronize with all of the GPUs on the system and wait for a long-running system call to complete.

To address these shortcomings, \texttt{torch-dgx} instead uses CUDA unified memory pools (via \texttt{cudaMallocFromPoolAsync}), which provide both stream-ordered memory management and memory reuse.
When an allocation is requested, we call \texttt{cudaMallocFromPoolAsync} on the non-default stream of a single device.
Then, CUDA events are used to synchronize the streams of all devices to the allocation stream to ensure that the allocation has completed before the memory is accessed.
For deallocation, we synchronize the streams of all devices to the deallocation stream to ensure that all devices have finished accessing the memory, then call \texttt{cudaFreeAsync} on the deallocation stream.
This asynchronous multi-GPU allocator eliminates the need for host synchronization in memory management.
Table~\ref{table:microbenchmarks} benchmarks representative distributed operators using this allocator, providing a baseline for evaluating its impact.
Table~\ref{table:mempool_ablation} repeats these benchmarks with a synchronous allocator without a memory pool.
Compared with Table~\ref{table:microbenchmarks}, the ablation shows substantial slowdowns for operators that allocate intermediate storage.

\begin{table}[t]
    \centering
    \caption{
        \texttt{torch-dgx} \emph{Memory Pool Ablation Study}:
        To assess the benefit of our asynchronous memory pool allocator, we replace it with a synchronous allocator that has no memory pool and therefore cannot reuse allocations.
        We report timings and relative speeds for the same GPU counts, operators, and problem sizes as those in Table~\ref{table:microbenchmarks}.
        The resulting host-device synchronization and repeated system calls substantially degrade performance.
        \label{table:mempool_ablation}}
    \scriptsize
    \setlength{\tabcolsep}{3pt}
    \begin{tabular*}{0.965\columnwidth}{@{\extracolsep{\fill}}l c cc cc cc@{}}
        \toprule
        & & \multicolumn{2}{c}{$K = 2^{26}$} & \multicolumn{2}{c}{$K = 2^{28}$} & \multicolumn{2}{c}{$K = 2^{30}$} \\
        \cmidrule(lr){3-4} \cmidrule(lr){5-6} \cmidrule(lr){7-8}
        Operator & GPUs & {\unit{\milli\second}} & {$\times$} & {\unit{\milli\second}} & {$\times$} & {\unit{\milli\second}} & {$\times$} \\
        \midrule
        \multirow{4}{*}{\texttt{fill}} & 1 & 0.1993 & 1.00 & 0.7812 & 1.00 & 3.1119 & 1.00 \\
        & 2 & 0.1167 & 1.71 & 0.4091 & 1.91 & 1.5744 & 1.98 \\
        & 4 & 0.0728 & 2.74 & 0.2191 & 3.57 & 0.8015 & 3.88 \\
        & 8 & 0.0769 & 2.59 & 0.1306 & 5.98 & 0.4223 & 7.37 \\
        \midrule
        \multirow{4}{*}{\texttt{add}} & 1 & 15.7184 & 1.00 & 61.4146 & 1.00 & 245.5594 & 1.00 \\
        & 2 & 14.0341 & 1.12 & 55.9619 & 1.10 & 223.3953 & 1.10 \\
        & 4 & 11.7256 & 1.34 & 46.8768 & 1.31 & 187.8200 & 1.31 \\
        & 8 & 10.5419 & 1.49 & 42.2731 & 1.45 & 168.3127 & 1.46 \\
        \midrule
        \multirow{4}{*}{\texttt{addmm}} & 1 & 76.47 & 1.00 & 542.84 & 1.00 & 4290.66 & 1.00 \\
        & 2 & 48.87 & 1.56 & 310.40 & 1.75 & 2180.44 & 1.97 \\
        & 4 & 36.59 & 2.09 & 201.81 & 2.69 & 1232.93 & 3.48 \\
        & 8 & 31.91 & 2.40 & 155.54 & 3.49 & 849.83 & 5.05 \\
        \midrule
        \multirow{4}{*}{\texttt{sum}} & 1 & 0.3692 & 1.00 & 0.7921 & 1.00 & 2.4841 & 1.00 \\
        & 2 & 0.4349 & 0.85 & 0.6943 & 1.14 & 1.4866 & 1.67 \\
        & 4 & 0.6036 & 0.61 & 0.7000 & 1.13 & 1.1259 & 2.21 \\
        & 8 & 1.0738 & 0.34 & 1.0844 & 0.73 & 1.2836 & 1.94 \\
        \midrule
        \multirow{4}{*}{\texttt{cumsum}} & 1 & 19.9854 & 1.00 & 78.7084 & 1.00 & 271.9306 & 1.00 \\
        & 2 & 15.4728 & 1.29 & 60.2653 & 1.31 & 238.7466 & 1.14 \\
        & 4 & 13.2441 & 1.51 & 52.0257 & 1.51 & 198.5987 & 1.37 \\
        & 8 & 12.1085 & 1.65 & 44.3939 & 1.77 & 174.6918 & 1.56 \\
        \bottomrule
    \end{tabular*}
\end{table}

\subsection{Distributed Operators}

Distributed PyTorch operators partition work across GPUs and execute their assigned chunks using concurrent CUDA kernels.
For \texttt{torch::add}, each GPU processes a disjoint range of elements, reading the corresponding inputs and writing their elementwise sums.
For \texttt{torch::sum}, each GPU processes a disjoint range of elements, computing a partial sum that is then combined with those from the other GPUs.
Unified memory allows these kernels to read both tensor elements and partial results directly across devices, without explicit staging or communication calls in model code.

Although unified memory makes data accessible across GPUs, execution speed still depends on maximizing the locality of each GPU's accesses.
For nonlocal accesses, CUDA unified memory uses a paging mechanism to migrate memory on demand between GPUs.
While this is sufficient for many cases, such as coherent reads across GPUs, it can sometimes be faster to prefetch inputs and outputs via \texttt{cudaMemPrefetchAsync} and \texttt{cudaMemPrefetchBatchAsync} if the memory segments accessed by a GPU are known (approximately) \emph{a priori}.
This trades the performance impact of page faults during kernel execution for prefetching overhead prior to kernel execution.
The performance characteristics often depend on the exact memory access pattern, and thus we perform prefetching judiciously in our implementation based on profiling.

After launching its kernels, each operator establishes dependencies between the GPU streams so that its kernels complete before those of the next operator begin.
These dependencies prevent cross-device read-write races when successive operators use different work partitions.
The host can therefore queue successive operators asynchronously, synchronizing with the GPUs only when results must be read back to the CPU.

Table~\ref{table:microbenchmarks} evaluates representative distributed operators on 1, 2, 4, and 8 NVLink-connected A100 GPUs.
\texttt{fill} and \texttt{add} exercise elementwise memory operations, \texttt{addmm} exercises matrix multiplication, and \texttt{sum} and \texttt{cumsum} exercise reductions and prefix sums.
At the largest problem size ($K=2^{30}$), these operators achieve speedups of $6.03$--$7.51\times$ on 8 GPUs relative to 1 GPU.
Smaller workloads provide less opportunity to amortize multi-GPU overhead, with \texttt{sum} at $K=2^{26}$ running slower on 8 GPUs than on 1 GPU.

Operator implementations can reuse existing CUDA libraries and optimization techniques.
We use CUB for sorting \cite{merrill2011sorting} and reductions \cite{merrill2016scan}, cuDNN for convolution \cite{chetlur2014cudnn}, and cuBLAS for matrix multiplication \cite{nvidia2025cublas}.
Conventional kernel optimizations also apply.
For example, we use vectorized loads and shared-memory caching of transcendental results in our fused Adam \cite{kingma2015adam} and AdamW implementations.

\subsection{PyTorch Integration and Ease of Use}

PyTorch dispatches each operator to an implementation keyed on the residency of its input tensors (e.g., \texttt{kCPU} corresponding to device \texttt{"cpu"}, \texttt{kCUDA} corresponding to device \texttt{"cuda"}).
The set of registered implementations for a key is called a backend.
We register our multi-GPU unified memory allocator and operator implementations under a new dispatch key, \texttt{kDGX}, corresponding to the device \texttt{"dgx"}.
Tensors allocated with \texttt{device="dgx"} then reside in unified memory, and operators with \texttt{"dgx"} inputs automatically dispatch to our multi-GPU implementations.
Automatic differentiation is handled either by specialized multi-GPU backward kernels or by decomposition through PyTorch autograd.

The supplementary material lists the currently supported PyTorch operators, which cover all operators invoked in the experiments described in Section~\ref{section:experiments}.

\section{Distributed Gaussian Splatting}\label{section:distributed_gaussian_splatting}

\begin{figure}[t]
    \centering
    \includegraphics[width=\columnwidth]{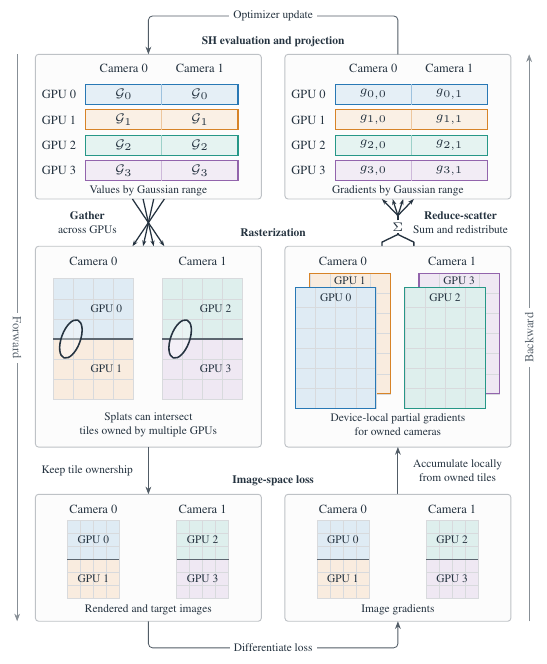}
    \caption{
        Distributed Gaussian splatting for $N=4$ GPUs and a batch of $B=2$ cameras. We preserve each operator's partitioning between the forward (left) and backward (right) passes to reuse data on the same GPUs.
        For spherical harmonic evaluation and projection, each GPU processes a disjoint range of Gaussians $\mathcal G_i$ across all cameras. In the backward passes, each Gaussian's owning GPU accumulates the cameras' gradient contributions $g_{i,c}$ using device-local atomics.
        For Gaussian-tile intersection, depth sorting, and rasterization, we partition output pixels into disjoint contiguous ranges of full-width tile rows. Unified memory provides access to projected Gaussians resident on other GPUs.
        A projected Gaussian may intersect tiles assigned to multiple GPUs, so the backward pass must combine gradient contributions computed on those GPUs.
        Each GPU accumulates its contributions in device-local gradient buffers (stacked grids), avoiding costly system-scope atomics.
        A reduce-scatter pass then sums these contributions and redistributes the resulting gradients to the Gaussian-owning GPUs, preserving locality for subsequent parameter updates.
        Image-space loss evaluation follows rasterization's tile ownership in both passes. We preserve this partition for rendered images, target images, and image gradients to reduce data movement.
        \label{figure:distributed_partitioning}}
\end{figure}

Gaussian splatting frameworks such as fVDB \cite{williams2024fvdb} and gsplat \cite{ye2025gsplat} combine core PyTorch operators with custom Gaussian splatting operators.
We distribute the core operators by changing the training device from \texttt{cuda} to \texttt{dgx}, which in turn places tensors in unified memory accessible to every GPU.
For the custom operators, we partition work to reduce data movement, page faults, and cross-GPU atomic contention.
As illustrated in Figure~\ref{figure:distributed_partitioning}, we preserve each operator's partitioning between its forward and backward passes to reuse data on the same GPUs.
Our implementation supports the general case of $N$ GPUs and a batch of $B$ cameras, with $B$ chosen independently of $N$.
It is not restricted to a single camera shared across all GPUs ($B=1$) or one camera per GPU ($B=N$), the latter being the configuration used by the gsplat \cite{ye2025gsplat} implementation of Grendel-GS \cite{zhao2025on}.

\begin{table}[t]
    \centering
    \caption{
        We compare the \texttt{torch-dgx} backend with PyTorch's built-in CUDA backend on a single A100 GPU with a batch size of 1. We report the Gaussian count, PSNR, SSIM, wall-clock time, and the speed of the \texttt{torch-dgx} backend relative to the CUDA backend. Consistent with the benchmarks in Tables~\ref{table:batch-scaling} and~\ref{table:tile-scaling}, the reconstructions ran for 200 training epochs on full-resolution images with 1/10th of the images held out for testing. The \texttt{torch-dgx} backend achieves similar reconstruction quality to the CUDA backend (Gaussian count, PSNR, and SSIM) while running at an average of $98.5\%$ of its speed.
        \label{table:backend-comparison}}
    \scriptsize
    \setlength{\tabcolsep}{4pt}
    \begin{tabular}{@{}l c c c c c c@{}}
        \toprule
        Scene & Backend & \# Splats & PSNR & SSIM & Time (\unit{\hour}) & Rel. speed \\
        \midrule
        \multirow{2}{*}{\shortstack[l]{\textsc{bonsai}\\$3118 \times 2078$}} & \texttt{cuda} & 5.407M & 32.879 & 0.9431 & 0.6014 & 1.000 \\
        & \texttt{dgx} & 5.434M & 32.613 & 0.9420 & 0.6181 & 0.973 \\
        \midrule
        \multirow{2}{*}{\shortstack[l]{\textsc{counter}\\$3115 \times 2076$}} & \texttt{cuda} & 6.280M & 29.505 & 0.9192 & 0.6158 & 1.000 \\
        & \texttt{dgx} & 6.323M & 29.091 & 0.9167 & 0.6247 & 0.986 \\
        \midrule
        \multirow{2}{*}{\shortstack[l]{\textsc{kitchen}\\$3115 \times 2078$}} & \texttt{cuda} & 8.203M & 32.169 & 0.9331 & 0.9469 & 1.000 \\
        & \texttt{dgx} & 8.230M & 32.021 & 0.9327 & 0.9597 & 0.987 \\
        \midrule
        \multirow{2}{*}{\shortstack[l]{\textsc{room}\\$3114 \times 2075$}} & \texttt{cuda} & 8.787M & 32.233 & 0.9281 & 0.7792 & 1.000 \\
        & \texttt{dgx} & 8.805M & 32.134 & 0.9258 & 0.7947 & 0.980 \\
        \midrule
        \multirow{2}{*}{\shortstack[l]{\textsc{stump}\\$4978 \times 3300$}} & \texttt{cuda} & 26.56M & 26.288 & 0.7886 & 0.6542 & 1.000 \\
        & \texttt{dgx} & 25.53M & 26.337 & 0.7884 & 0.6589 & 0.993 \\
        \midrule
        \multirow{2}{*}{\shortstack[l]{\textsc{garden}\\$5187 \times 3361$}} & \texttt{cuda} & 39.99M & 26.282 & 0.7996 & 1.613 & 1.000 \\
        & \texttt{dgx} & 40.07M & 26.347 & 0.7999 & 1.631 & 0.989 \\
        \midrule
        \multirow{2}{*}{\shortstack[l]{\textsc{bicycle}\\$4946 \times 3286$}} & \texttt{cuda} & 36.02M & 24.217 & 0.7209 & 1.274 & 1.000 \\
        & \texttt{dgx} & 36.12M & 24.228 & 0.7214 & 1.286 & 0.991 \\
        \bottomrule
    \end{tabular}
\end{table}

For spherical harmonic evaluation \cite{kerbl2023gaussian} and Gaussian projection, each GPU processes a disjoint range of Gaussians across all cameras in the batch.
These operators produce view-dependent colors and projected geometry, respectively.
We compute SH view directions directly from Gaussian means and camera matrices within the kernel, avoiding a separate direction tensor and its associated memory traffic.
In the backward passes, each Gaussian's owning GPU accumulates gradient contributions from all cameras using device-local atomics.
This avoids the additional overhead of cross-GPU atomics.

\begin{table*}[t]
    \centering
    \caption{
        \emph{Increasing Batch Size Evaluation}. Using the Mip-NeRF 360 dataset \cite{barron2022mipnerf} (MIT license), we evaluate our multi-GPU Gaussian splatting approach alongside the Gaussian splatting approach in gsplat \cite{ye2025gsplat}, which implements Grendel-GS \cite{zhao2025on}, on $N \in \{1, 2, 4, 8\}$ A100 GPUs connected with NVLink.
        For each scene, we train on the full-resolution images, holding out 1/10 for testing.
        The training batch size $B$ is equal to the number of GPUs, and we report Gaussian count, PSNR \cite{huynhthu2008psnr}, SSIM, wall-clock time, and relative speed over 200 epochs.
        Since our implementation and gsplat use different hyperparameters and Gaussian densification strategies, we tune gsplat's refinement parameters to obtain Gaussian counts roughly comparable to those of our approach.
        Even with this tuning, gsplat often has difficulty densifying to the higher counts reached by our implementation.
        Our approach achieves mean speedups of $1.68\times$, $2.50\times$, and $3.15\times$ at 2, 4, and 8 GPUs and a peak speedup of $3.60\times$, while gsplat achieves mean speedups of $1.77\times$, $3.33\times$, and $6.30\times$ and a peak speedup of $7.30\times$.
        Speedups are relative to each implementation's own single-GPU baseline.
        \label{table:batch-scaling}}
    \scriptsize
    \setlength{\tabcolsep}{4pt}
    \begin{tabular*}{\textwidth}{@{}l c @{\extracolsep{\fill}} ccccc ccccc@{}}
        \toprule
        & & \multicolumn{5}{c}{\texttt{torch-dgx} (ours)} & \multicolumn{5}{c}{gsplat \cite{ye2025gsplat}} \\
        \cmidrule(lr){3-7} \cmidrule(lr){8-12}
        Scene & $N{=}B$ & \# Splats & PSNR & SSIM & Time (\unit{\hour}) & Rel. speed & \# Splats & PSNR & SSIM & Time (\unit{\hour}) & Rel. speed \\
        \midrule
            \multirow{4}{*}{\shortstack[l]{\textsc{bonsai}\\$3118 \times 2078$}} & 1 & 5.434M & 32.613 & 0.9420 & 0.6181 & 1.000 & 6.502M & 32.154 & 0.9399 & 0.9697 & 1.000 \\
             & 2 & 5.386M & 33.003 & 0.9430 & 0.3811 & 1.622 & 6.321M & 32.641 & 0.9410 & 0.5665 & 1.712 \\
             & 4 & 5.393M & 32.961 & 0.9426 & 0.2575 & 2.400 & 6.287M & 32.034 & 0.9400 & 0.2814 & 3.446 \\
             & 8 & 5.467M & 32.452 & 0.9396 & 0.2044 & 3.023 & 6.244M & 32.118 & 0.9372 & 0.1476 & 6.571 \\
            \midrule
            \multirow{4}{*}{\shortstack[l]{\textsc{counter}\\$3115 \times 2076$}} & 1 & 6.323M & 29.091 & 0.9167 & 0.6247 & 1.000 & 5.800M & 28.712 & 0.9093 & 0.8350 & 1.000 \\
             & 2 & 6.256M & 29.355 & 0.9187 & 0.3811 & 1.639 & 5.877M & 28.589 & 0.9020 & 0.4366 & 1.913 \\
             & 4 & 6.215M & 29.444 & 0.9190 & 0.2542 & 2.458 & 5.799M & 28.794 & 0.9161 & 0.2495 & 3.346 \\
             & 8 & 6.148M & 29.378 & 0.9173 & 0.1958 & 3.190 & 5.695M & 28.794 & 0.9143 & 0.1340 & 6.231 \\
            \midrule
            \multirow{4}{*}{\shortstack[l]{\textsc{kitchen}\\$3115 \times 2078$}} & 1 & 8.230M & 32.021 & 0.9327 & 0.9597 & 1.000 & 8.253M & 32.115 & 0.9332 & 1.1148 & 1.000 \\
             & 2 & 8.096M & 32.154 & 0.9330 & 0.5536 & 1.734 & 8.151M & 32.111 & 0.9341 & 0.6052 & 1.842 \\
             & 4 & 8.082M & 32.061 & 0.9330 & 0.3550 & 2.703 & 8.339M & 31.606 & 0.9309 & 0.2888 & 3.860 \\
             & 8 & 8.249M & 31.720 & 0.9315 & 0.2664 & 3.603 & 8.419M & 31.775 & 0.9340 & 0.1711 & 6.515 \\
            \midrule
            \multirow{4}{*}{\shortstack[l]{\textsc{room}\\$3114 \times 2075$}} & 1 & 8.805M & 32.134 & 0.9258 & 0.7947 & 1.000 & 7.387M & 30.451 & 0.9057 & 1.0189 & 1.000 \\
             & 2 & 8.605M & 32.373 & 0.9281 & 0.4789 & 1.660 & 7.665M & 31.463 & 0.9207 & 0.5881 & 1.732 \\
             & 4 & 8.366M & 32.399 & 0.9287 & 0.3131 & 2.539 & 7.400M & 29.933 & 0.9078 & 0.3280 & 3.107 \\
             & 8 & 8.245M & 32.150 & 0.9277 & 0.2358 & 3.370 & 7.082M & 29.671 & 0.8959 & 0.1747 & 5.831 \\
            \midrule
            \multirow{4}{*}{\shortstack[l]{\textsc{stump}\\$4978 \times 3300$}} & 1 & 25.53M & 26.337 & 0.7884 & 0.6589 & 1.000 & 16.153M & 26.363 & 0.7981 & 0.8557 & 1.000 \\
             & 2 & 25.22M & 26.285 & 0.7892 & 0.4117 & 1.601 & 16.650M & 26.342 & 0.7991 & 0.5326 & 1.607 \\
             & 4 & 26.52M & 26.307 & 0.7895 & 0.3100 & 2.125 & 17.939M & 26.254 & 0.7977 & 0.3056 & 2.800 \\
             & 8 & 28.04M & 26.237 & 0.7892 & 0.2808 & 2.346 & 17.915M & 26.234 & 0.7966 & 0.1644 & 5.204 \\
            \midrule
            \multirow{4}{*}{\shortstack[l]{\textsc{garden}\\$5187 \times 3361$}} & 1 & 40.07M & 26.347 & 0.7999 & 1.631 & 1.000 & 31.837M & 26.271 & 0.8052 & 2.1121 & 1.000 \\
             & 2 & 39.51M & 26.050 & 0.7993 & 0.9017 & 1.809 & 32.229M & 26.329 & 0.8070 & 1.1373 & 1.857 \\
             & 4 & 38.90M & 26.331 & 0.8002 & 0.6031 & 2.705 & 32.136M & 26.172 & 0.8039 & 0.5852 & 3.609 \\
             & 8 & 38.16M & 26.279 & 0.7997 & 0.4706 & 3.467 & 32.233M & 26.122 & 0.8051 & 0.2893 & 7.300 \\
            \midrule
            \multirow{4}{*}{\shortstack[l]{\textsc{bicycle}\\$4946 \times 3286$}} & 1 & 36.12M & 24.228 & 0.7214 & 1.286 & 1.000 & 24.884M & 24.117 & 0.7226 & 1.7194 & 1.000 \\
             & 2 & 35.89M & 24.204 & 0.7212 & 0.7492 & 1.716 & 25.978M & 24.074 & 0.7215 & 0.9934 & 1.731 \\
             & 4 & 35.83M & 24.225 & 0.7214 & 0.5039 & 2.551 & 27.081M & 24.121 & 0.7227 & 0.5485 & 3.135 \\
             & 8 & 35.30M & 24.201 & 0.7210 & 0.4186 & 3.071 & 26.605M & 23.995 & 0.7211 & 0.2659 & 6.466 \\
        \bottomrule
    \end{tabular*}
\end{table*}

For the dense image rendering used in our experiments, we configure Gaussian rasterization to use $16\times16$ pixel tiles, matching the fixed tile size of the image-space loss operators.
We partition the output pixels across all cameras into disjoint contiguous ranges, with each range comprising one or more full-width rows of tiles.
To establish depth order for alpha blending, each GPU first examines all projected Gaussians for cameras whose images contain its assigned tiles.
Although some projected Gaussians may reside on other GPUs, CUDA unified memory provides access through a shared address space, with data transferred over the GPU interconnect (e.g., NVLink or PCIe).
Following AccuTile in Speedy-Splat \cite{hanson2025speedysplat}, we use tight ellipse-based opacity isocontour tile culling to reduce the number of intersection records allocated, sorted, and traversed.
Each GPU computes intersections with its assigned tiles and radix sorts the resulting Gaussian-tile pairs by camera, tile, and depth.
To further reduce storage, we scan intersection counts per block of Gaussians and reconstruct per-Gaussian offsets when emitting the intersection records.
Because the tile ranges are disjoint and ordered, concatenating the independently sorted results produces a globally sorted sequence without a cross-GPU merge.

In the forward rasterization pass, each GPU rasterizes the sorted projected Gaussians intersecting its assigned tiles.
A projected Gaussian may intersect tiles assigned to multiple GPUs, so the backward pass must combine gradient contributions computed on those GPUs.
System-scope atomics would require cross-GPU coordination for scattered gradient updates, increasing communication overhead and limiting scaling.
Instead, each GPU accumulates its contributions in device-local gradient buffers.
A reduce-scatter pass then sums these contributions and redistributes the resulting gradients to the Gaussian-owning GPUs.

Image-space loss evaluation follows rasterization's tile ownership in both the forward and backward passes.
The forward pass sums tile contributions locally, then reduces across GPUs to obtain the scalar loss.
The backward pass computes image gradients for each GPU's assigned tiles, with halo reads handled through unified memory.
We preserve this partition when copying target images to the GPUs and converting between interleaved and planar layouts, reducing data movement across these image-space operations.
We also extend fused SSIM \cite{wang2004ssim, mallick2024taming} to compute a weighted combination of L1 and SSIM losses, reducing global-memory accesses and intermediate tensor allocations.

Together with the distributed PyTorch backend, these custom operators form a complete multi-GPU Gaussian splatting pipeline.
We next evaluate its training performance, reconstruction quality, and capacity for large-scale scenes.


\section{Experiments} \label{section:experiments}

\begin{figure*}[t]
    \centering
    \includegraphics[width=0.66\linewidth]{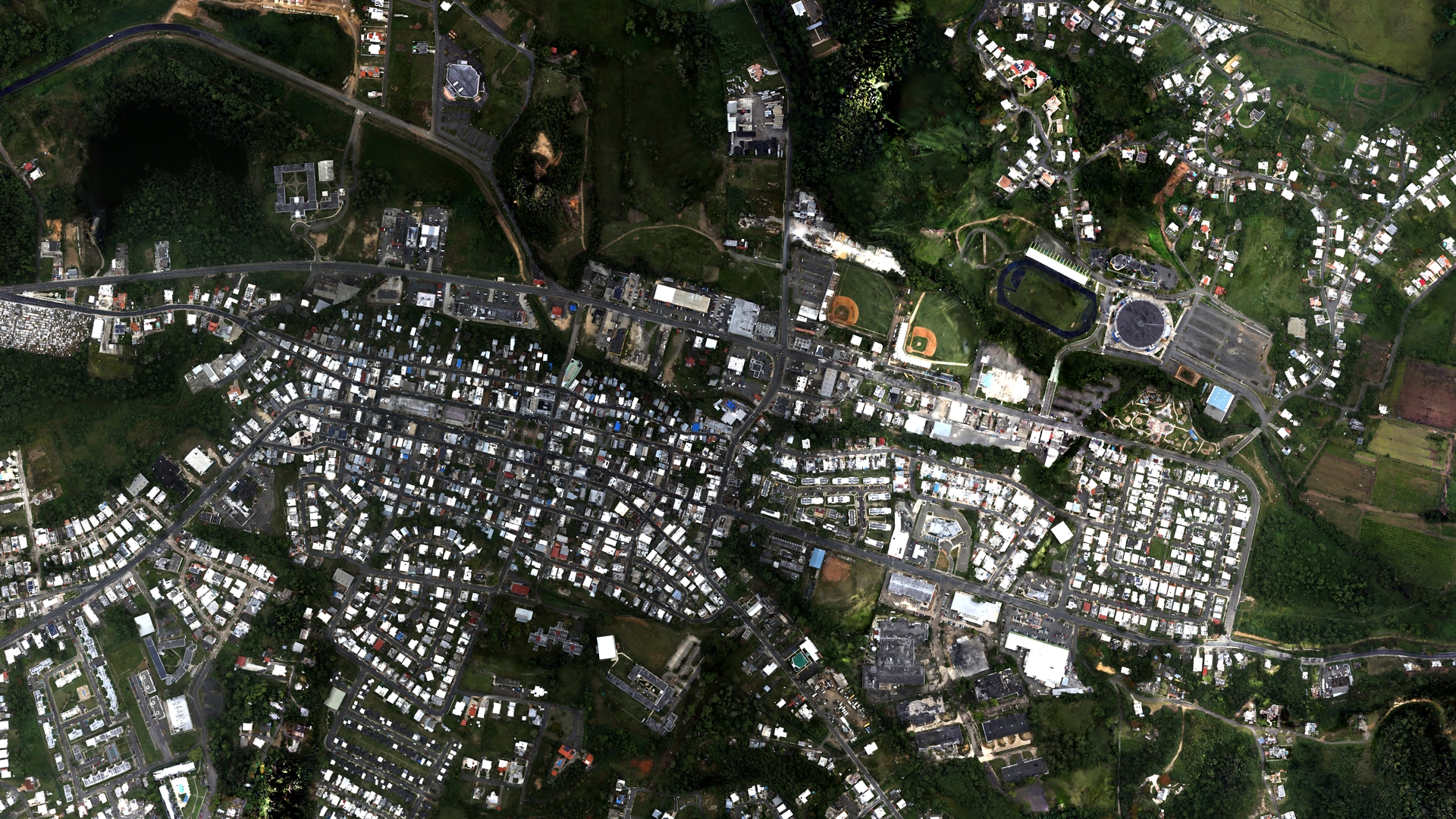}
    \includegraphics[width=0.33\linewidth]{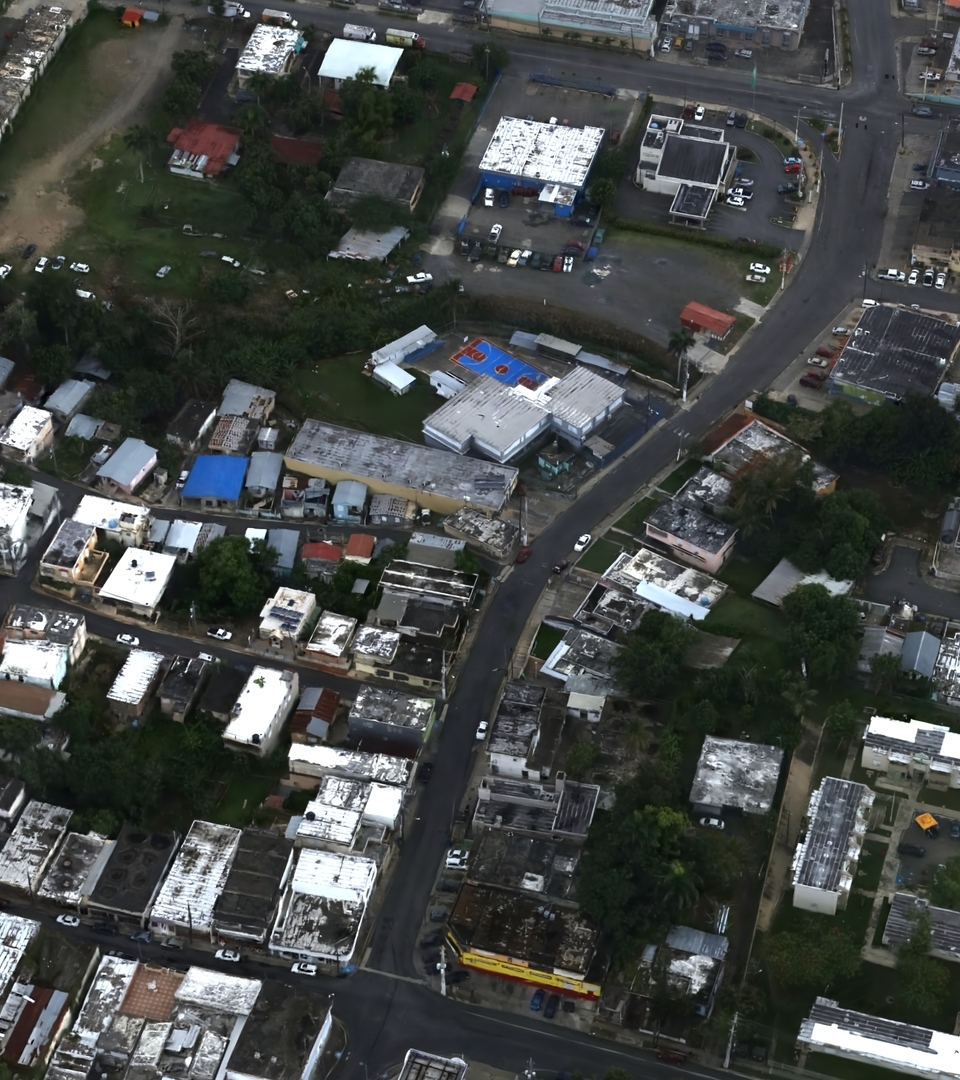}
    \caption{
        A stress test of our approach: a city-scale high-resolution reconstruction pushed to over 1 billion Gaussian splats trained using a batch size of 1.
        The dataset consists of 1850 aerial images at $8807 \times 5825$ resolution covering 7 square kilometers, with camera poses initialized via COLMAP.
        Each training step requires rasterizing the Gaussian splats at the resolution of the dataset, maximizing the detail in the final reconstruction at the cost of significantly higher memory and bandwidth pressure.
        In fact, the resulting Gaussian data alone (excluding gradients and intermediate quantities) far exceeds the memory of any single GPU, so even the views shown here are produced via distributed rendering.
        This reconstruction was trained progressively on 8 A100s, then 8 H100s, then 4 B200s without needing to modify model code or checkpoints, exemplifying the portability of our approach.
        \emph{Left:}~overhead camera view showing the sheer scale of the reconstruction.
        The reconstruction covers multiple city blocks and contains both man-made and natural features at a variety of scales.
        \emph{Right:}~zoomed-in camera view demonstrating the high degree of detail attainable at this scale.
        Although the reconstruction spans several city blocks, the rendering still resolves minute street-level details such as windshields, lane markings, and lampposts.
        This is made possible by densifying to an extremely large Gaussian count alongside training on high-resolution images.
        Dataset courtesy of the Civil Air Patrol.
        \label{figure:puerto_rico_reconstruction}}
\end{figure*}

We evaluate the single-GPU overhead of \texttt{torch-dgx} and multi-GPU Gaussian splatting with increasing and constant batch sizes on the well-established Mip-NeRF 360 \cite{barron2022mipnerf} dataset.
We then demonstrate the reconstruction capacity of our approach on large aerial datasets.
All built-in PyTorch operators used in the \texttt{torch-dgx} training runs reported in Tables~\ref{table:backend-comparison}--\ref{table:tile-scaling} and shown in Figures~\ref{figure:gettysburg_reconstruction} and~\ref{figure:puerto_rico_reconstruction} dispatch to our multi-GPU implementations.
Section~\ref{section:backend} evaluates the underlying backend through operator benchmarks and an allocator ablation (Tables~\ref{table:microbenchmarks} and~\ref{table:mempool_ablation}).

\subsection{Gaussian Splatting Benchmarks}

Table~\ref{table:backend-comparison} compares the \texttt{torch-dgx} and built-in PyTorch CUDA backends for Gaussian splat training on one A100 GPU with $B=1$.
Developed in a short period, \texttt{torch-dgx} achieves similar reconstruction quality and runs at an average of $98.5\%$ of the speed of PyTorch's CUDA backend, which has benefited from years of community development and optimization.

For the experiments in Tables~\ref{table:backend-comparison}--\ref{table:tile-scaling}, we use seven scenes from the Mip-NeRF 360 \cite{barron2022mipnerf} dataset.
We train for 200 epochs on full-resolution images, holding out every tenth image for testing.
For the multi-GPU experiments, we use $N \in \{1, 2, 4, 8\}$ A100 GPUs connected with NVLink, with $B=N$ in Table~\ref{table:batch-scaling} and $B=1$ in Table~\ref{table:tile-scaling}.
Increasing $B$ therefore reduces the number of optimizer updates over the 200 epochs.

In Table~\ref{table:batch-scaling}, we increase the batch size with the number of GPUs, training on multiple cameras simultaneously.
Following \cite{zhao2025on}, we scale learning rates by $\sqrt{B}$ and set Adam's momentum coefficients to $(0.9^B, 0.999^B)$ while also dividing Adam's numerical-stability parameter $\epsilon$ by $\sqrt{B}$.
Since our implementation and gsplat use different hyperparameters and Gaussian densification strategies, we tune gsplat's refinement parameters to obtain Gaussian counts roughly comparable to those of our approach.
Even with this tuning, gsplat often has difficulty densifying to the higher counts reached by our implementation.

Across the dataset, we obtain mean speedups of $1.68\times$, $2.50\times$, and $3.15\times$ at 2, 4, and 8 GPUs, respectively, relative to our single-GPU baseline.
The corresponding mean speedups for gsplat are $1.77\times$, $3.33\times$, and $6.30\times$, relative to its own single-GPU baseline.
In five of the seven single-GPU runs, our approach completes training in less time while producing more Gaussians than gsplat.
Our approach also demonstrates broadly consistent Gaussian counts and reconstruction quality across batch sizes.
Increasing the batch size from 1 to 8 reduces mean PSNR by 0.051 dB and mean SSIM by 0.0001 after 200 epochs.
In contrast, the gsplat results vary more on some scenes, especially for \textsc{room}, where gsplat's PSNR decreases from 31.463 dB at 2 GPUs to 29.671 dB at 8 GPUs.

\begin{table}[t]
    \centering
    \centering
    \caption{
        \emph{Constant Batch Size Evaluation}. Using the Mip-NeRF 360 dataset \cite{barron2022mipnerf} (MIT license), we evaluate our multi-GPU Gaussian splatting approach on $N \in \{1, 2, 4, 8\}$ A100 GPUs connected with NVLink.
        For each scene, we train on the full-resolution images, holding out 1/10 for testing.
        The training batch size $B$ is held constant at 1 across GPU counts, and we report Gaussian count, PSNR, SSIM, wall-clock time, and relative speed over 200 epochs.
        The implementation of Grendel-GS \cite{zhao2025on} in gsplat \cite{ye2025gsplat} does not support $B=1$ with multiple GPUs ($N>1$).
        Gaussian counts, PSNR, and SSIM remain broadly consistent across GPU counts for each scene.
        With $B=1$, our approach achieves mean speedups across scenes of $1.44\times$, $1.85\times$, and $2.18\times$ at 2, 4, and 8 GPUs, respectively, relative to one GPU.
        Scaling from 4 to 8 GPUs yields diminishing returns; scattered cross-device accesses during rasterization are a likely contributor.
        \label{table:tile-scaling}}
    \scriptsize
    \setlength{\tabcolsep}{4pt}
    \begin{tabular}{@{}l c c c c c c c@{}}
        \toprule
        Scene & $N$ & $B$ & \# Splats & PSNR & SSIM & Time (\unit{\hour}) & Rel. speed \\
        \midrule
        \multirow{4}{*}{\shortstack[l]{\textsc{bonsai}\\$3118 \times 2078$}} & 1 & 1 & 5.434M & 32.613 & 0.9420 & 0.6181 & 1.000 \\
        & 2 & 1 & 5.441M & 32.822 & 0.9407 & 0.4400 & 1.405 \\
        & 4 & 1 & 5.411M & 33.044 & 0.9435 & 0.3506 & 1.763 \\
        & 8 & 1 & 5.416M & 32.846 & 0.9396 & 0.3253 & 1.900 \\
        \midrule
        \multirow{4}{*}{\shortstack[l]{\textsc{counter}\\$3115 \times 2076$}} & 1 & 1 & 6.323M & 29.091 & 0.9167 & 0.6247 & 1.000 \\
        & 2 & 1 & 6.335M & 29.385 & 0.9187 & 0.4506 & 1.387 \\
        & 4 & 1 & 6.300M & 29.330 & 0.9181 & 0.3547 & 1.761 \\
        & 8 & 1 & 6.322M & 29.232 & 0.9180 & 0.3078 & 2.030 \\
        \midrule
        \multirow{4}{*}{\shortstack[l]{\textsc{kitchen}\\$3115 \times 2078$}} & 1 & 1 & 8.230M & 32.021 & 0.9327 & 0.9597 & 1.000 \\
        & 2 & 1 & 8.265M & 31.981 & 0.9330 & 0.6447 & 1.489 \\
        & 4 & 1 & 8.150M & 32.100 & 0.9329 & 0.5017 & 1.913 \\
        & 8 & 1 & 8.201M & 32.053 & 0.9329 & 0.4106 & 2.338 \\
        \midrule
        \multirow{4}{*}{\shortstack[l]{\textsc{room}\\$3114 \times 2075$}} & 1 & 1 & 8.805M & 32.134 & 0.9258 & 0.7947 & 1.000 \\
        & 2 & 1 & 8.842M & 32.373 & 0.9290 & 0.5542 & 1.434 \\
        & 4 & 1 & 8.759M & 32.289 & 0.9284 & 0.4300 & 1.848 \\
        & 8 & 1 & 8.669M & 32.329 & 0.9286 & 0.3881 & 2.048 \\
        \midrule
        \multirow{4}{*}{\shortstack[l]{\textsc{stump}\\$4978 \times 3300$}} & 1 & 1 & 25.53M & 26.337 & 0.7884 & 0.6589 & 1.000 \\
        & 2 & 1 & 24.93M & 26.271 & 0.7888 & 0.4725 & 1.394 \\
        & 4 & 1 & 25.95M & 26.298 & 0.7890 & 0.3747 & 1.758 \\
        & 8 & 1 & 24.73M & 26.274 & 0.7887 & 0.3114 & 2.116 \\
        \midrule
        \multirow{4}{*}{\shortstack[l]{\textsc{garden}\\$5187 \times 3361$}} & 1 & 1 & 40.07M & 26.347 & 0.7999 & 1.631 & 1.000 \\
        & 2 & 1 & 40.08M & 26.316 & 0.7999 & 1.074 & 1.519 \\
        & 4 & 1 & 40.02M & 26.272 & 0.7999 & 0.7822 & 2.086 \\
        & 8 & 1 & 40.08M & 26.284 & 0.7994 & 0.6264 & 2.604 \\
        \midrule
        \multirow{4}{*}{\shortstack[l]{\textsc{bicycle}\\$4946 \times 3286$}} & 1 & 1 & 36.12M & 24.228 & 0.7214 & 1.286 & 1.000 \\
        & 2 & 1 & 35.95M & 24.232 & 0.7217 & 0.8892 & 1.446 \\
        & 4 & 1 & 36.27M & 24.242 & 0.7215 & 0.7106 & 1.809 \\
        & 8 & 1 & 36.50M & 24.210 & 0.7209 & 0.5808 & 2.213 \\
        \bottomrule
    \end{tabular}
\end{table}

For this experiment, our scaling is likely limited by cross-GPU communication and workload imbalance.
Even though the $B=N$ case decouples per-view rasterization, the 3D Gaussians are still distributed across GPUs.
Peer reads occur when a GPU rasterizes a view using projected outputs computed on another GPU.
Uneven rendering workloads can also leave some GPUs idle while others finish.
Dynamic workload rebalancing \cite{zhao2025on} could mitigate this effect.

In Table~\ref{table:tile-scaling}, we hold the batch size constant at 1 and rasterize disjoint image tiles on each GPU.
Our approach distributes both model parameters and projected Gaussian data across GPUs, allowing a single camera view to use their aggregate memory, which increases linearly with $N$.
Gaussian counts, PSNR, and SSIM remain broadly consistent across GPU counts, and all multi-GPU runs are faster than the single-GPU baseline.
Mean speedups across scenes are $1.44\times$, $1.85\times$, and $2.18\times$ at 2, 4, and 8 GPUs, respectively.
With these modest Gaussian counts, scaling from 4 to 8 GPUs yields diminishing returns, providing additional speedups of 8--25\% across scenes (18\% on average).

In batched training, a camera's projected Gaussian data can reside on the GPU that rasterizes its image.
With $B=1$, multiple GPUs render tiles of the same image and may need projected Gaussian data resident on other devices.
Scattered accesses over the GPU interconnect are therefore a likely contributor to the diminishing speedups in Table~\ref{table:tile-scaling}.
To improve the spatial locality of distributed rasterization, we Morton sort the Gaussians in three-dimensional space \cite{morton1966computer, liao2025litegs} during densification.
Morton sorting provides a 5--10\% performance improvement, although three-dimensional locality does not guarantee locality in image space for a given camera.

\subsection{Large-Scale Reconstruction}

Figure~\ref{figure:gettysburg_reconstruction} demonstrates our approach on a Gettysburg reconstruction with over 268 million Gaussian splats from aerial photography spanning 13 square kilometers.
We train with $B=1$ for 400 epochs on 8 A100 GPUs in 3.086 hours, obtaining a PSNR of 27.450 dB and an SSIM of 0.8734 on held-out images.
For comparison, we run the implementation of Grendel-GS \cite{zhao2025on} in gsplat \cite{ye2025gsplat} on 8 A100 GPUs with $B=8$ and run out of memory at merely 132 million Gaussians.
This is the minimum supported batch size for this GPU count, as the gsplat implementation distributes only the 3D Gaussian parameters across GPUs, whereas each view's projected parameters must remain on the single GPU that rasterizes that view.
Our approach also distributes these per-view parameters, allowing a single view to use the aggregate GPU memory.
On the same eight A100 GPUs, our completed reconstruction contains over twice the Gaussian count gsplat reached before an out-of-memory crash.

To probe the limits of our approach, we stress-test it with a city-scale reconstruction of over 1 billion Gaussian splats (Figure~\ref{figure:puerto_rico_reconstruction}), more than 25 times the largest Gaussian count reported in \cite{zhao2025on}.
We train this reconstruction progressively on 8 A100s, then 8 H100s, then 4 B200s, without modifying model code or checkpoints.
Even though the source data is from aerial photography, our reconstruction is able to resolve street-level details.
Both reconstructions far exceed a single GPU's rendering memory capacity, so we render them using the distributed forward pass.

\section{Conclusion}

We demonstrate city-scale reconstructions with street-level detail and over 1 billion Gaussian splats.
We distribute Gaussian parameters and per-view rendering data so each view can use multiple GPUs' aggregate memory and compute.
At the core is \texttt{torch-dgx}, a PyTorch backend that exposes multiple GPUs as an aggregate device.
Built on CUDA unified memory with optional NVLink support, it combines a stream-ordered memory pool with distributed operators to keep allocation and execution asynchronous with respect to the host.
For models composed of supported operators, a device-argument change distributes execution without explicit communication or sharding annotations in model code.
Beyond Gaussian splatting, this operator-level design supports other PyTorch workloads.
By placing multi-GPU support in reusable operators, \texttt{torch-dgx} allows new applications to build on existing distributed implementations and share future optimizations.
We have open-sourced \texttt{torch-dgx} and its evaluated operators as a foundation for community development.

%

{
    \small
    \bibliographystyle{ieeenat_fullname}
    \bibliography{references}
}

\clearpage
\appendix
\maketitlesupplementary

\section{Gaussian Splatting Variants}\label{sec:gsplat-variants}

To demonstrate the generality of our approach to other Gaussian splatting workloads, we implement a multi-GPU version of the base tile-sorted variant of 3DGUT \cite{wu20253dgut}, which supports distorted cameras through unscented-transform (UT) projection and world-space rasterization.
This variant reuses the backend's distributed PyTorch operators and the custom spherical harmonic evaluation, Gaussian-tile intersection, and fused loss operators described in Section~\ref{section:distributed_gaussian_splatting}.

\begin{table}[t]
    \centering
    \caption{
        \emph{3DGUT Increasing Batch Size Evaluation}. Using the Mip-NeRF 360 dataset \cite{barron2022mipnerf} (MIT license), we evaluate our multi-GPU 3DGUT implementation \cite{wu20253dgut} on $N \in \{1, 2, 4, 8\}$ A100 GPUs connected with NVLink.
        For each scene, we train on the full-resolution images, holding out 1/10 for testing.
        The training batch size $B$ is equal to the number of GPUs, and we report Gaussian count, PSNR \cite{huynhthu2008psnr}, SSIM, wall-clock time, and relative speed over 200 epochs.
        Our implementation achieves mean speedups of $1.66\times$, $2.70\times$, and $4.19\times$ at 2, 4, and 8 GPUs and a peak speedup of $5.31\times$.
        Speedups are relative to the single-GPU baseline.
        \label{table:3dgut-batch-scaling}}
    \scriptsize
    \setlength{\tabcolsep}{4pt}
    \begin{tabular}{@{}l c c c c c c@{}}
        \toprule
        Scene & $N{=}B$ & \# Splats & PSNR & SSIM & Time (\unit{\hour}) & Rel. speed \\
        \midrule
        \multirow{4}{*}{\shortstack[l]{\textsc{bonsai}\\$3118 \times 2078$}} & 1 & 9.357M & 32.978 & 0.9447 & 2.306 & 1.000 \\
         & 2 & 9.357M & 32.823 & 0.9443 & 1.302 & 1.770 \\
         & 4 & 9.637M & 32.812 & 0.9445 & 0.8203 & 2.811 \\
         & 8 & 10.53M & 32.733 & 0.9443 & 0.4847 & 4.757 \\
            \midrule
        \multirow{4}{*}{\shortstack[l]{\textsc{counter}\\$3115 \times 2076$}} & 1 & 7.054M & 29.606 & 0.9227 & 2.315 & 1.000 \\
         & 2 & 7.054M & 29.673 & 0.9230 & 1.314 & 1.762 \\
         & 4 & 7.266M & 29.663 & 0.9234 & 0.7606 & 3.044 \\
         & 8 & 7.266M & 29.568 & 0.9229 & 0.4442 & 5.212 \\
            \midrule
        \multirow{4}{*}{\shortstack[l]{\textsc{kitchen}\\$3115 \times 2078$}} & 1 & 10.93M & 32.361 & 0.9364 & 3.039 & 1.000 \\
         & 2 & 11.26M & 32.327 & 0.9356 & 1.761 & 1.726 \\
         & 4 & 11.26M & 32.335 & 0.9356 & 0.9803 & 3.101 \\
         & 8 & 10.93M & 31.899 & 0.9333 & 0.5722 & 5.312 \\
            \midrule
        \multirow{4}{*}{\shortstack[l]{\textsc{room}\\$3114 \times 2075$}} & 1 & 5.100M & 32.870 & 0.9315 & 1.906 & 1.000 \\
         & 2 & 5.100M & 32.880 & 0.9315 & 1.184 & 1.610 \\
         & 4 & 5.253M & 32.859 & 0.9316 & 0.7156 & 2.664 \\
         & 8 & 5.253M & 32.907 & 0.9319 & 0.4297 & 4.436 \\
            \midrule
        \multirow{4}{*}{\shortstack[l]{\textsc{stump}\\$4978 \times 3300$}} & 1 & 1.451M & 26.346 & 0.8089 & 0.7519 & 1.000 \\
         & 2 & 1.451M & 26.198 & 0.8072 & 0.5203 & 1.445 \\
         & 4 & 1.451M & 26.264 & 0.8072 & 0.3594 & 2.092 \\
         & 8 & 1.732M & 26.119 & 0.8058 & 0.2978 & 2.525 \\
            \midrule
        \multirow{4}{*}{\shortstack[l]{\textsc{garden}\\$5187 \times 3361$}} & 1 & 6.284M & 26.548 & 0.8122 & 1.890 & 1.000 \\
         & 2 & 6.284M & 26.559 & 0.8127 & 1.124 & 1.681 \\
         & 4 & 6.667M & 26.576 & 0.8131 & 0.6997 & 2.701 \\
         & 8 & 7.503M & 26.574 & 0.8133 & 0.5178 & 3.650 \\
            \midrule
        \multirow{4}{*}{\shortstack[l]{\textsc{bicycle}\\$4946 \times 3286$}} & 1 & 2.457M & 24.347 & 0.7310 & 1.530 & 1.000 \\
         & 2 & 2.457M & 24.365 & 0.7311 & 0.9464 & 1.617 \\
         & 4 & 2.457M & 24.239 & 0.7301 & 0.6167 & 2.481 \\
         & 8 & 2.766M & 24.314 & 0.7317 & 0.4406 & 3.473 \\
        \bottomrule
    \end{tabular}
\end{table}

\begin{table}[t]
    \centering
    \caption{
        \emph{3DGUT Constant Batch Size Evaluation}. Using the Mip-NeRF 360 dataset \cite{barron2022mipnerf} (MIT license), we evaluate our multi-GPU 3DGUT implementation \cite{wu20253dgut} on $N \in \{1, 2, 4, 8\}$ A100 GPUs connected with NVLink.
        For each scene, we train on the full-resolution images, holding out 1/10 for testing.
        The training batch size $B$ is held constant at 1 across GPU counts, and we report Gaussian count, PSNR, SSIM, wall-clock time, and relative speed over 200 epochs.
        Gaussian counts, PSNR, and SSIM remain broadly consistent across GPU counts for each scene.
        With $B=1$, our approach achieves mean speedups across scenes of $1.48\times$, $2.12\times$, and $3.04\times$ at 2, 4, and 8 GPUs, respectively, relative to one GPU.
        \label{table:3dgut-tile-scaling}}
    \scriptsize
    \setlength{\tabcolsep}{4pt}
    \begin{tabular}{@{}l c c c c c c c@{}}
        \toprule
        Scene & $N$ & $B$ & \# Splats & PSNR & SSIM & Time (\unit{\hour}) & Rel. speed \\
        \midrule
        \multirow{4}{*}{\shortstack[l]{\textsc{bonsai}\\$3118 \times 2078$}} & 1 & 1 & 9.357M & 32.978 & 0.9447 & 2.306 & 1.000 \\
        & 2 & 1 & 9.357M & 32.894 & 0.9436 & 1.551 & 1.486 \\
        & 4 & 1 & 9.357M & 32.940 & 0.9448 & 1.053 & 2.190 \\
        & 8 & 1 & 9.357M & 32.968 & 0.9445 & 0.8111 & 2.843 \\
        \midrule
        \multirow{4}{*}{\shortstack[l]{\textsc{counter}\\$3115 \times 2076$}} & 1 & 1 & 7.054M & 29.606 & 0.9227 & 2.315 & 1.000 \\
        & 2 & 1 & 7.054M & 29.674 & 0.9227 & 1.569 & 1.475 \\
        & 4 & 1 & 7.054M & 29.602 & 0.9226 & 1.035 & 2.237 \\
        & 8 & 1 & 7.054M & 29.566 & 0.9228 & 0.7075 & 3.272 \\
        \midrule
        \multirow{4}{*}{\shortstack[l]{\textsc{kitchen}\\$3115 \times 2078$}} & 1 & 1 & 10.93M & 32.361 & 0.9364 & 3.039 & 1.000 \\
        & 2 & 1 & 10.93M & 32.327 & 0.9353 & 2.001 & 1.519 \\
        & 4 & 1 & 10.93M & 32.385 & 0.9357 & 1.556 & 1.953 \\
        & 8 & 1 & 10.93M & 32.274 & 0.9354 & 1.014 & 2.996 \\
        \midrule
        \multirow{4}{*}{\shortstack[l]{\textsc{room}\\$3114 \times 2075$}} & 1 & 1 & 5.100M & 32.870 & 0.9315 & 1.906 & 1.000 \\
        & 2 & 1 & 5.100M & 32.848 & 0.9317 & 1.242 & 1.535 \\
        & 4 & 1 & 5.100M & 32.856 & 0.9315 & 0.8244 & 2.312 \\
        & 8 & 1 & 5.100M & 32.794 & 0.9314 & 0.5925 & 3.217 \\
        \midrule
        \multirow{4}{*}{\shortstack[l]{\textsc{stump}\\$4978 \times 3300$}} & 1 & 1 & 1.451M & 26.346 & 0.8089 & 0.7519 & 1.000 \\
        & 2 & 1 & 1.451M & 26.319 & 0.8086 & 0.5353 & 1.405 \\
        & 4 & 1 & 1.451M & 26.308 & 0.8087 & 0.3922 & 1.917 \\
        & 8 & 1 & 1.451M & 26.231 & 0.8080 & 0.2892 & 2.600 \\
        \midrule
        \multirow{4}{*}{\shortstack[l]{\textsc{garden}\\$5187 \times 3361$}} & 1 & 1 & 6.284M & 26.548 & 0.8122 & 1.890 & 1.000 \\
        & 2 & 1 & 6.284M & 26.521 & 0.8120 & 1.228 & 1.539 \\
        & 4 & 1 & 6.284M & 26.549 & 0.8121 & 0.8347 & 2.264 \\
        & 8 & 1 & 6.284M & 26.518 & 0.8122 & 0.5469 & 3.455 \\
        \midrule
        \multirow{4}{*}{\shortstack[l]{\textsc{bicycle}\\$4946 \times 3286$}} & 1 & 1 & 2.457M & 24.347 & 0.7310 & 1.530 & 1.000 \\
        & 2 & 1 & 2.457M & 24.313 & 0.7309 & 1.086 & 1.409 \\
        & 4 & 1 & 2.457M & 24.294 & 0.7313 & 0.7694 & 1.988 \\
        & 8 & 1 & 2.457M & 24.364 & 0.7310 & 0.5250 & 2.914 \\
        \bottomrule
    \end{tabular}
\end{table}

For UT projection, each GPU processes a disjoint range of Gaussians across all cameras in the batch, following the partition used for standard Gaussian projection.
This operator estimates projected footprints for tile culling.
For world-space rasterization, we partition output pixels into disjoint contiguous ranges of full-width tile rows.
In the forward pass, each GPU evaluates the contributing Gaussians at their points of maximum response along camera rays and alpha blends their contributions.
In the backward pass, the rasterizer computes gradients directly with respect to 3D geometry, color, and opacity.

\begin{figure}[t]
    \centering
    \includegraphics[width=0.94\columnwidth]{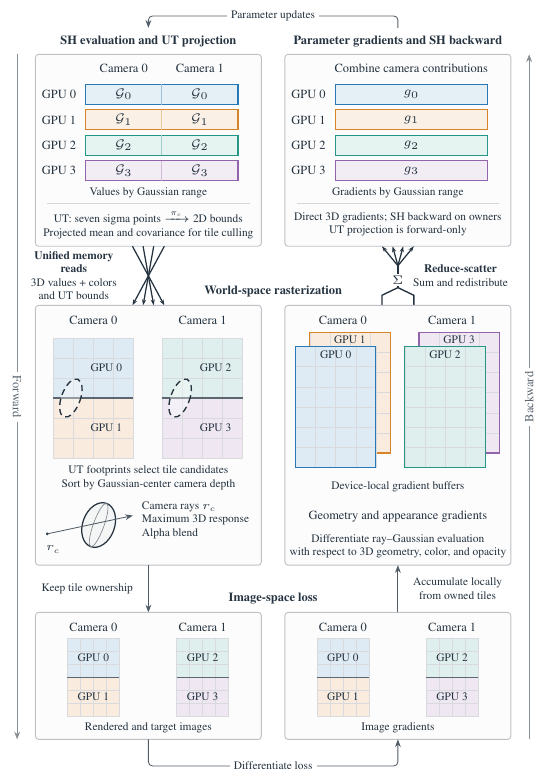}
    \caption{
        Distributed 3DGUT for $N=4$ GPUs and a batch of $B=2$ cameras. We preserve Gaussian and tile ownership between the forward (left) and backward (right) passes.
        For spherical harmonic evaluation and unscented-transform (UT) projection, each GPU processes a disjoint range of Gaussians $\mathcal G_i$ across all cameras. Dashed ellipses denote UT footprints for tile culling.
        For Gaussian-tile intersection, depth sorting, and world-space rasterization, we partition output pixels into disjoint contiguous ranges of full-width tile rows. Unified memory provides access to projected bounds and 3D Gaussian parameters resident on other GPUs.
        The rasterizer computes geometry gradients directly; UT projection is forward-only.
        Each GPU accumulates gradients for Gaussian geometry and appearance in device-local buffers (stacked grids), avoiding system-scope atomics.
        A reduce-scatter pass then sums contributions across GPUs and redistributes the resulting gradients to the Gaussian-owning GPUs, where upstream backward passes complete the parameter gradients $g_i$.
        Image-space loss evaluation follows rasterization's tile ownership in both passes.
        \label{figure:distributed_partitioning_3dgut}}
\end{figure}

A Gaussian may contribute to tiles assigned to multiple GPUs, so the backward pass must combine gradient contributions computed on those GPUs.
Each GPU accumulates its contributions in device-local gradient buffers, avoiding system-scope atomics.
A reduce-scatter pass then sums these contributions and redistributes the resulting gradients to the Gaussian-owning GPUs for upstream backward passes.
Figure~\ref{figure:distributed_partitioning_3dgut} illustrates the full multi-GPU training loop.

Using the partitioning described in Section~\ref{section:distributed_gaussian_splatting}, we evaluate 3DGUT with increasing and constant batch sizes, as in our standard Gaussian splatting experiments.
In Table~\ref{table:3dgut-batch-scaling}, we increase the batch size with the number of GPUs ($B=N$), so each GPU rasterizes one camera view.
On eight GPUs, training is $4.19\times$ faster on average than on one GPU, with the largest speedup on \textsc{kitchen} ($5.31\times$).
Across the seven scenes, increasing the batch size from one to eight reduces mean PSNR by $0.135$~dB and mean SSIM by $0.0006$.
In Table~\ref{table:3dgut-tile-scaling}, we keep $B=1$ and distribute the tile rows of a single view across GPUs.
On eight GPUs, training is $3.04\times$ faster on average than on one GPU, and PSNR differs from its single-GPU value by less than $0.12$~dB in every scene.
Because 3DGUT and standard Gaussian splatting have similar communication patterns, we expect cross-GPU communication and workload imbalance to limit scaling in both. When GPUs share a view, scattered reads of remote Gaussian data may slow rasterization.

\section{Beyond Gaussian Splatting}\label{sec:diffusion-demo}

To exercise \texttt{torch-dgx} on a workload outside three-dimensional reconstruction, we run the sampling step of a diffusion probabilistic model \cite{ho2020denoising}.
We run the implementation from \cite{esser2020pytorchdiffusion} through the \texttt{kDGX} backend by changing the model's device from \texttt{cuda} to \texttt{dgx}.
This model invokes a different set of operators from those invoked by Gaussian splatting and further exercises the operator inventory of Table~\ref{table:operator-list}.
We have also trained a diffusion transformer (DiT) \cite{peebles2023scalable} using \texttt{torch-dgx}, demonstrating its applicability to neural network training beyond Gaussian splatting.
Every operator invoked during diffusion sampling and DiT training dispatches to \texttt{kDGX} with no CPU fallback, and we verify numerical correctness by comparing the \texttt{dgx} backend against the \texttt{cuda} reference.

\section{Ease of Use}\label{sec:ease-of-use}

Listing~\ref{listing:code_sample} provides an illustrative, self-contained training example using \texttt{torch-dgx}.
The model and optimizer use the existing PyTorch APIs.
Changing the device to \texttt{dgx} places tensors in unified memory and distributes operators, enabling the model to train on multiple GPUs.

\section{Operator Coverage}

Table~\ref{table:operator-list} lists the PyTorch operators currently implemented in \texttt{torch-dgx}.
These cover all built-in operations used by our Gaussian splatting pipelines and the diffusion sampling experiment in Section~\ref{sec:diffusion-demo}.
The backend also supports neural-network training through forward and backward computations for common layers, normalization, attention, and losses, together with parameter updates.
Unsupported operators can fall back to PyTorch's CPU implementation with a user warning.
Users can then run the workload end to end and see which operators lack distributed implementations.

Individual operators can be implemented independently and tested against PyTorch reference implementations, providing a practical workflow for LLM-assisted development.
We used LLM assistance to implement many of the distributed operators.

\begin{listing}
    \renewcommand{\theFancyVerbLine}{\textcolor{gray}{\rmfamily\tiny\arabic{FancyVerbLine}}}
\begin{minted}{Python}
import torch, numpy, torch_dgx
device = "dgx:0"  # usually "cpu" or "cuda:0"
model = torch.nn.Sequential(torch.nn.Linear(1, 64),
                            torch.nn.ReLU(),
                            torch.nn.Linear(64, 1)).to(device)
x_train = numpy.random.uniform(
    -3, 3, size=(100, 1)).astype("float32")
y_train = x_train**2
x_train = torch.tensor(x_train, device=device)
y_train = torch.tensor(y_train, device=device)
loss_fn = torch.nn.MSELoss()
optimizer = torch.optim.Adam(model.parameters(),
                             lr=0.1, fused=True)
for i in range(10):
    outputs = model(x_train)
    loss = loss_fn(outputs, y_train)
    print(f"Iteration {i} Loss = {loss.item()}")
    optimizer.zero_grad()
    loss.backward()
    optimizer.step()
\end{minted}
    \caption{Training a simple model with \texttt{torch-dgx}.
    Distributing an existing model only requires changing the device from \texttt{cuda} (or \texttt{cpu}) to \texttt{dgx} (line 2).
    Tensors then reside in unified memory and operators are parallelized across devices automatically.
    Within the training loop, the host synchronizes with the GPUs only when \texttt{loss.item()} reads back the loss on line 17.\label{listing:code_sample}}
\end{listing}

\begin{table}[t]
    \centering
    \captionsetup{font=footnotesize}
    \caption{
        Supported PyTorch operations in \texttt{torch-dgx}. During the backward pass, PyTorch autograd evaluates derivative formulas that can combine existing tensor operations or invoke dedicated multi-GPU backward kernels.
        \label{table:operator-list}}
    \fontsize{7}{7.8}\selectfont
    \renewcommand{\arraystretch}{1.0}
    \setlength{\tabcolsep}{0pt}
    \begin{tabularx}{\columnwidth}{@{}
        >{\raggedright\arraybackslash}p{0.18\columnwidth}@{\hspace{6pt}}
        >{\raggedright\fontfamily{zi4}\selectfont\arraybackslash}X@{}}
        \toprule
        \textbf{Category} & {\normalfont\bfseries Operator families} \\
        \midrule
        Math and activations &
            \mbox{abs},
            \mbox{add},
            \mbox{addcdiv},
            \mbox{addcmul},
            \mbox{atan2},
            \mbox{ceil},
            \mbox{clamp},
            \mbox{clamp\_max},
            \mbox{clamp\_min},
            \mbox{cos},
            \mbox{div},
            \mbox{exp},
            \mbox{floor},
            \mbox{floor\_divide},
            \mbox{floor\_divide\_},
            \mbox{gelu},
            \mbox{hypot},
            \mbox{lerp},
            \mbox{log},
            \mbox{log10},
            \mbox{logit},
            \mbox{minimum},
            \mbox{mul},
            \mbox{nan\_to\_num},
            \mbox{neg},
            \mbox{polar},
            \mbox{pow},
            \mbox{reciprocal},
            \mbox{relu},
            \mbox{relu\_},
            \mbox{remainder},
            \mbox{round},
            \mbox{rsqrt},
            \mbox{sgn},
            \mbox{sigmoid},
            \mbox{sign},
            \mbox{silu},
            \mbox{sin},
            \mbox{sqrt},
            \mbox{sub},
            \mbox{tanh},
            \mbox{threshold\_backward} \\
        \addlinespace[2.5pt]
        Comparisons and logic &
            \mbox{bitwise\_and},
            \mbox{bitwise\_not},
            \mbox{bitwise\_or},
            \mbox{eq},
            \mbox{equal},
            \mbox{ge},
            \mbox{gt},
            \mbox{isnan},
            \mbox{isneginf},
            \mbox{le},
            \mbox{logical\_and},
            \mbox{logical\_or},
            \mbox{lt},
            \mbox{ne} \\
        \addlinespace[2.5pt]
        Reductions and scans &
            \mbox{all},
            \mbox{amax},
            \mbox{any},
            \mbox{argmax},
            \mbox{argmin},
            \mbox{bincount},
            \mbox{cumprod},
            \mbox{cumsum},
            \mbox{linalg\_vector\_norm},
            \mbox{max},
            \mbox{mean},
            \mbox{median},
            \mbox{min},
            \mbox{prod},
            \mbox{sum},
            \mbox{var\_mean} \\
        \addlinespace[2.5pt]
        Linear algebra &
            \mbox{addmm},
            \mbox{bmm},
            \mbox{linalg\_cross},
            \mbox{linalg\_inv\_ex},
            \mbox{\_linalg\_solve\_ex},
            \mbox{mm},
            \mbox{tril},
            \mbox{triu} \\
        \addlinespace[2.5pt]
        Indexing and selection &
            \mbox{gather},
            \mbox{index},
            \mbox{index\_add},
            \mbox{index\_fill\_},
            \mbox{\_index\_put\_impl\_},
            \mbox{index\_select},
            \mbox{masked\_fill\_},
            \mbox{masked\_select},
            \mbox{nonzero},
            \mbox{scatter},
            \mbox{scatter\_add},
            \mbox{sort},
            \mbox{topk},
            \mbox{where} \\
        \addlinespace[2.5pt]
        Tensor construction &
            \mbox{arange},
            \mbox{cat},
            \mbox{eye},
            \mbox{fill\_},
            \mbox{flip},
            \mbox{repeat\_interleave},
            \mbox{zero\_} \\
        \addlinespace[2.5pt]
        Random sampling &
            \mbox{multinomial},
            \mbox{normal\_},
            \mbox{random\_},
            \mbox{randperm},
            \mbox{uniform\_} \\
        \addlinespace[2.5pt]
        Layers, normalization, and losses &
            \mbox{avg\_pool2d},
            \mbox{convolution\_overrideable},
            \mbox{embedding\_dense\_backward},
            \mbox{\_fused\_rms\_norm},
            \mbox{\_fused\_sdp\_choice},
            \mbox{grid\_sampler\_2d},
            \mbox{linear},
            \mbox{\_log\_softmax},
            \mbox{mse\_loss},
            \mbox{native\_batch\_norm},
            \mbox{native\_dropout},
            \mbox{native\_layer\_norm\_backward},
            \mbox{nll\_loss2d},
            \mbox{nll\_loss},
            \mbox{reflection\_pad2d},
            \mbox{\_safe\_softmax},
            \mbox{\_scaled\_dot\_product\_efficient\_attention},
            \mbox{\_softmax},
            \mbox{upsample\_nearest2d} \\
        \addlinespace[2.5pt]
        Optimizers &
            \mbox{\_fused\_adam\_},
            \mbox{\_fused\_adamw\_};
            \mbox{\_foreach\_*} families: add, sub, mul, div, maximum, lerp,
            addcmul, addcdiv, sqrt, reciprocal, neg
            {\normalfont (in-place and out-of-place variants)} \\
        \midrule
        Allocation, views, and storage &
            \mbox{as\_strided},
            \mbox{\_copy\_from},
            \mbox{\_copy\_from\_and\_resize},
            \mbox{empty},
            \mbox{empty\_strided},
            \mbox{\_local\_scalar\_dense},
            \mbox{\_reshape\_alias},
            \mbox{resize\_},
            \mbox{set\_},
            \mbox{unfold},
            \mbox{view},
            \mbox{view\_as\_complex},
            \mbox{view\_as\_real} \\
        \bottomrule
    \end{tabularx}
\end{table}

%

\end{document}